\documentclass{article}
\usepackage{spconf,amsmath,graphicx}

\usepackage{enumitem}
\setlist{nosep, leftmargin=14pt}
\usepackage[verbose]{placeins}
\usepackage{mathtools}
\usepackage{framed,multirow}
\usepackage{booktabs}
\usepackage{float}
\usepackage{amssymb}
\usepackage{hyperref}
\DeclarePairedDelimiter\norm{\lVert}{\rVert}%

\title{Self-Supervised Modality-Agnostic Pre-Training of Swin Transformers}
\name{Abhiroop Talasila, Maitreya Maity, U. Deva Priyakumar \thanks{Corresponding author: deva@iiit.ac.in}}
\address{Center for Computational Natural Sciences and Bioinformatics\\
International Institute of Information Technology\\
Hyderabad, India\\}
\begin{document}
%
\maketitle
\begin{abstract}
Unsupervised pre-training has emerged as a transformative paradigm, displaying remarkable advancements in various domains. However, the susceptibility to domain shift, where pre-training data distribution differs from fine-tuning, poses a significant obstacle. To address this, we augment the Swin Transformer to learn from different medical imaging modalities, enhancing downstream performance. Our model, dubbed SwinFUSE (Swin Multi-Modal Fusion for UnSupervised Enhancement), offers three key advantages: (i) it learns from both Computed Tomography (CT) and Magnetic Resonance Images (MRI) during pre-training, resulting in complementary feature representations; (ii) a domain-invariance module (DIM) that effectively highlights salient input regions, enhancing adaptability; (iii) exhibits remarkable generalizability, surpassing the confines of tasks it was initially pre-trained on.
Our experiments on two publicly available 3D segmentation datasets show a modest 1-2\% performance trade-off compared to single-modality models, yet significant out-performance of up to 27\% on out-of-distribution modality. This substantial improvement underscores our proposed approach's practical relevance and real-world applicability. Code is available at: \href{https://github.com/devalab/SwinFUSE}{https://github.com/devalab/SwinFUSE}

\end{abstract}


\section{Introduction}
\label{sec:intro}

Supervised deep learning excels at medical image segmentation using lots of labeled data \cite{dou20173d, xu2022omega, truong2021transferable}. The shortage of professional radiologists and their limited time and annotation efficiency makes it difficult to get huge medical picture collections with exact annotations. Thus, routine clinical usage of supervised-learning-based segmentation techniques is limited. Recent research has focused on self-supervised learning (SSL), which uses many unlabeled images to learn the general aspects of medical images. Fully supervised model fine-tuning uses a minimal quantity of labeled data \cite{zhuang2019self}. Effective self-supervised medical image segmentation depends on pre-training quality. 

Contrastive learning is a self-supervised pre-training method that minimizes the latent space distance of pairs of similar images (typically produced from the same original image using different data augmentation processes) and maximizes the distance of pairs of dissimilar ones \cite{chen2020big}. Contrastive learning methods address domain shift, enhancing model applicability in downstream segmentation networks through consistent data augmentation strategies ensuring similar input distributions. Existing self-supervised learning has two limitations \cite{zhang2023multi}.

\begin{figure}[]
    \includegraphics[width=\columnwidth]{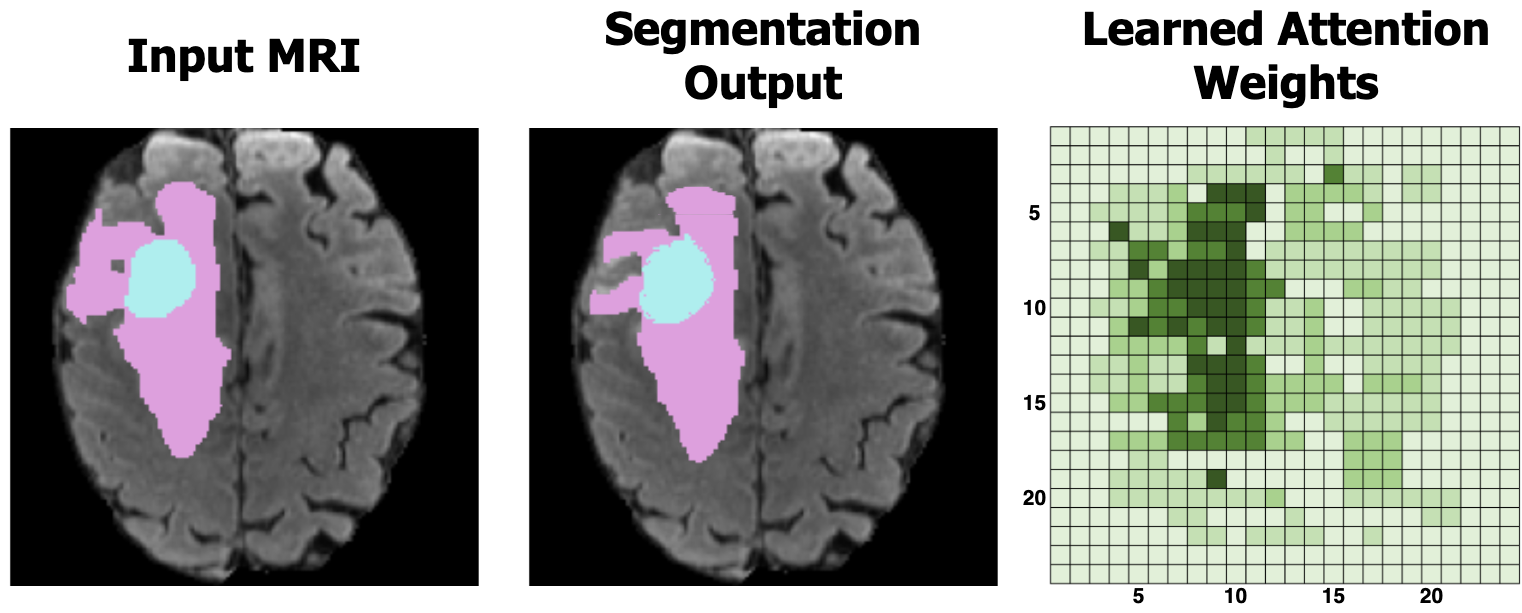} 
    \caption{Visual interpretation of SwinFUSE's attention weights (darker shades indicate higher relevance) for a BraTS21 MRI and the model's segmentation output.}
\label{fig:maps} 
\end{figure}

\begin{itemize}
    \item Domain Shift: Upstream pre-training uses modified images, affecting downstream segmentation network input distributions. General features from pre-trained models may not apply to segmentation networks.
    \item Multi-modality: Current techniques often rely on single-modal data, missing the benefits of multiple modalities. Multi-modal images offer diverse perspectives and augment network segmentation information.
\end{itemize}

\begin{figure*}[ht]
\centering
\includegraphics[width=0.9\textwidth]{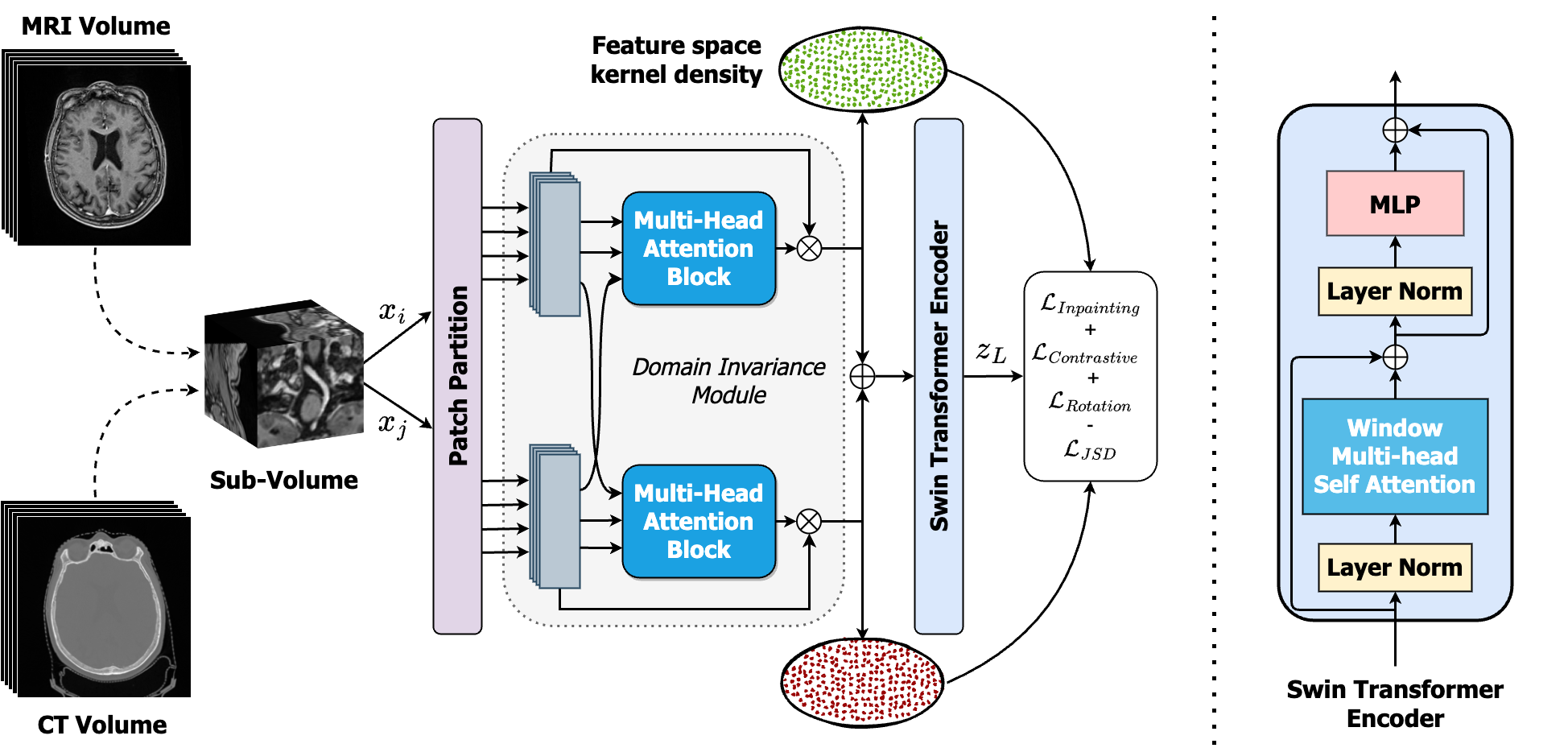} 
\caption{Outline of our proposed pre-training pipeline. Sub-volumes are randomly created from input images and augmented with random inner cutouts and rotations ($x_{i}, x_{j}$). Each augmentation passes through the patch partition layer to generate embeddings, which are fed to the DIM. The output from the DIM is extracted as kernel densities and forwarded to the Swin Transformer.}
\label{fig:arch} 
\end{figure*}

Vision Transformers (ViTs) transformed medical image analysis and computer vision. Transformers thrive in pre-text tasks, large-scale training, and layer-based global and local knowledge learning. ViTs simulate long-range global information using self-attention blocks and encode visual representations from patches, unlike Convolutional Neural Networks (CNNs) with small receptive fields. A hierarchical ViT with Shifted Windows (Swin) for local self-attention computing with non-overlapping windows was developed by Liu et al. \cite{liu2021swin}. Linear architecture has been found to be more efficient than ViT's quadratic self-attention layers. Swin UNETR \cite{hatamizadeh2021swin} merges feature maps at various sizes using transformer-encoded spatial representations in convolution-based decoders and achieves state-of-the-art (SOTA) performance in BTCV multi-organ segmentation and Medical Segmentation Decathlon (MSD) challenges \cite{antonelli2022medical}. The training paradigm uses proxy activities to learn human anatomical patterns. We extend this intuition to allow complementary feature learning from multiple imaging modalities.

This paper presents SwinFUSE, a modality-invariant self-supervised pre-training approach for medical image analysis. We utilize Swin UNETR's contrastive learning, masked volume pinpointing, and 3D rotation prediction as proxy tasks for pre-training. Additionally, we introduce a DIM for concurrent feature learning from CT and MRI data. The DIM implicitly identifies relevant input areas and directs them to the Swin Transformer encoder using attention maps (Fig. \ref{fig:maps}). We train the network on the SynthRad dataset \cite{synthrad} and retain the DIM and encoder for later fine-tuning. Our 3D image segmentation experiments involve the BraTS21 \cite{baid2021rsna} and MSD datasets. We fine-tune the entire network to demonstrate generalization across domains, validating its efficacy for each task, including organ segmentation in the MSD dataset.


\section{Method}
\label{sec:methods}

\subsection{Datasets}
\label{ssec:datasets}

\paragraph*{SynthRAD}{comprises registered brain and pelvis CT images with cone-beam CT and MRI images, serving the purpose of synthetic CT generation for radiotherapy planning \cite{synthrad}. We focus on a subset of 180 patients, utilizing T1-weighted gradient-echo MRIs, with some using contrast.}

\vspace{-10px}

\paragraph*{BraTS21}{consists of multi-modal MRI scans of glioma, with a total of 1254 patients \cite{baid2021rsna}. The sequences acquired include T1, T2, T1CE, and FLAIR. Segmentation classes include peritumoral edematous/invaded tissue, tumor core, and necrotic tumor core.}

\vspace{-10px}

\paragraph*{MSD}{contains 2,633 3D images collected from various anatomical regions, modalities, and medical image sources for segmentation purposes \cite{antonelli2022medical}. It covers data on body organs or parts like the Brain, Heart, Liver, Lung, Pancreas, Prostate, Hepatic Vessel, Hippocampus, Spleen, and Colon.}

\subsection{Pre-training}
\label{ssec:pt}

We augment the Swin UNETR architecture using a novel Domain Invariance Module, trained to learn which features to highlight, conditioned on the input type. The training dataset consists of CTs and MRIs; volumes are sampled randomly. During each iteration, 3D patches measuring $x_{n} \in \mathbb{R}^{96 \times 96 \times 96}$ voxels undergo augmentation with random inner cutout and rotation. These patches are then projected into a C-dimensional space ($C = 48$) using an embedding layer leading to the DIM as shown in Fig. \ref{fig:arch}. The two embeddings from respective augmentation are each fed into a 4-layer deep Multi-Head Attention Block (MHA) with $3, 6, 12, 24$ heads respectively like Co-Attention \cite{siam2020weakly}. The $Q, K, V$ embedding dimensions increase by an exponent of $2$ with the base layer having dimensions $x \in \mathbb{R}^{48}$. The query vector from the first embedding is fed as query input to the second MHA block and vice versa. Each block in the DIM is initialized with an embedding dimension of $2304$. The attention weights are scaled with the original embeddings, and the resulting average embedding is sent as input to the Swin Transformer Encoder. The DIM is constructed as given below, where $P$ denotes the patch partitions being fed into each MHA block ($\Psi$).

\vspace{-14px}

\begin{equation}
    DIM: P_1 \cdot \Psi_1(Q_2, K_1, V_1) + \\
    P_2 \cdot \Psi_2(Q_1, K_2, V_2)
    \label{}
\end{equation}

We train the model on the SynthRad dataset using the AdamW optimizer and a warm-up cosine scheduler with 500 iterations on two RTX 3090's. We set the initial learning rate for the pre-training experiments to $4e^{-4}$ and a decay of $1e^{-5}$. We implement our model using PyTorch and MONAI \footnote{https://monai.io/}.

\subsection{Loss Function}
\label{ssec:lf}

We aim to minimize the loss of Swin UNETR's encoder using multiple pre-training objectives, including masked volume inpainting, 3D image rotation, and contrastive coding. Additionally, we maximize an extra loss term that, akin to the approach in \cite{xie2022unsupervised}, employs non-parametric density estimation through kernel density estimation (KDE) and density matching via Jenson-Shannon divergence (JSD). This density-matching loss is a regularizer, ensuring that the feature distribution overlap between the source and target datasets is minimized. The KDE, denoted as $p_{est}(X)$, is formulated as follows:

\begin{equation}
    p_{est}(X) = \frac{1}{N} \sum_{n=1}^{N} K \left(\frac{\norm{X - X_{n}}_2}{\sigma}\right)
\end{equation}

where $X_1, X_2, X_3, \cdots, X_N$ is the number of sampled points from the encoded feature space, the output from the MHA block in our model, and $K$ is a Gaussian kernel. The bandwidth parameter ($\sigma$) is estimated to be the mean of the distance between the nearest neighbors in the feature space.

Our loss term for density matching, $\mathcal{L}_{\text{JSD}}$, given density of each MHA block outputs as $p_1$ and $p_2$, is given as follows:

\vspace{-10px}
\begin{equation}
    JSD_{p_1,p_2} = \frac{1}{2} \{KL[p_1, M]+ KL[p_2, M]\} 
\end{equation}

where KL is the KL divergence between the two distributions and $M$ is the average of both density estimates. The final loss:

\vspace{-10px}
\begin{equation}
\mathcal{L}_{\text{total}}=\mathcal{L}_{\text{inpaint}}+\mathcal{L}_{\text{contrast}} +\mathcal{L}_{\text{rot}}-\mathcal{L}_{\text{JSD}}
\end{equation}

\subsection{Fine-tuning}
\label{ssec:ft}

In the downstream task, such as 3D image segmentation, we fine-tune the complete Swin UNETR model by removing the projection heads while retaining the DIM. During training, sub-volumes are randomly cropped from the volumetric data. Then, stochastic data augmentations, including random rotation and cutout, are applied twice to each sub-volume within a mini-batch, resulting in two different views of each data. All other augmentation parameters align with those used in Swin UNETR.

For SwinFUSE, we utilize pre-trained weights for both the CT and MRI tasks, following the official methods outlined in nnUnet \cite{isensee2019nnu} and Swin UNETR \cite{hatamizadeh2021swin}. To ensure robustness, we employ a five-fold cross-validation strategy to train models for BraTS21 and MSD experiments. In each fold, we select the best model and ensemble their outputs to generate the final segmentation predictions.

\begin{figure}[]
    \includegraphics[width=\columnwidth]{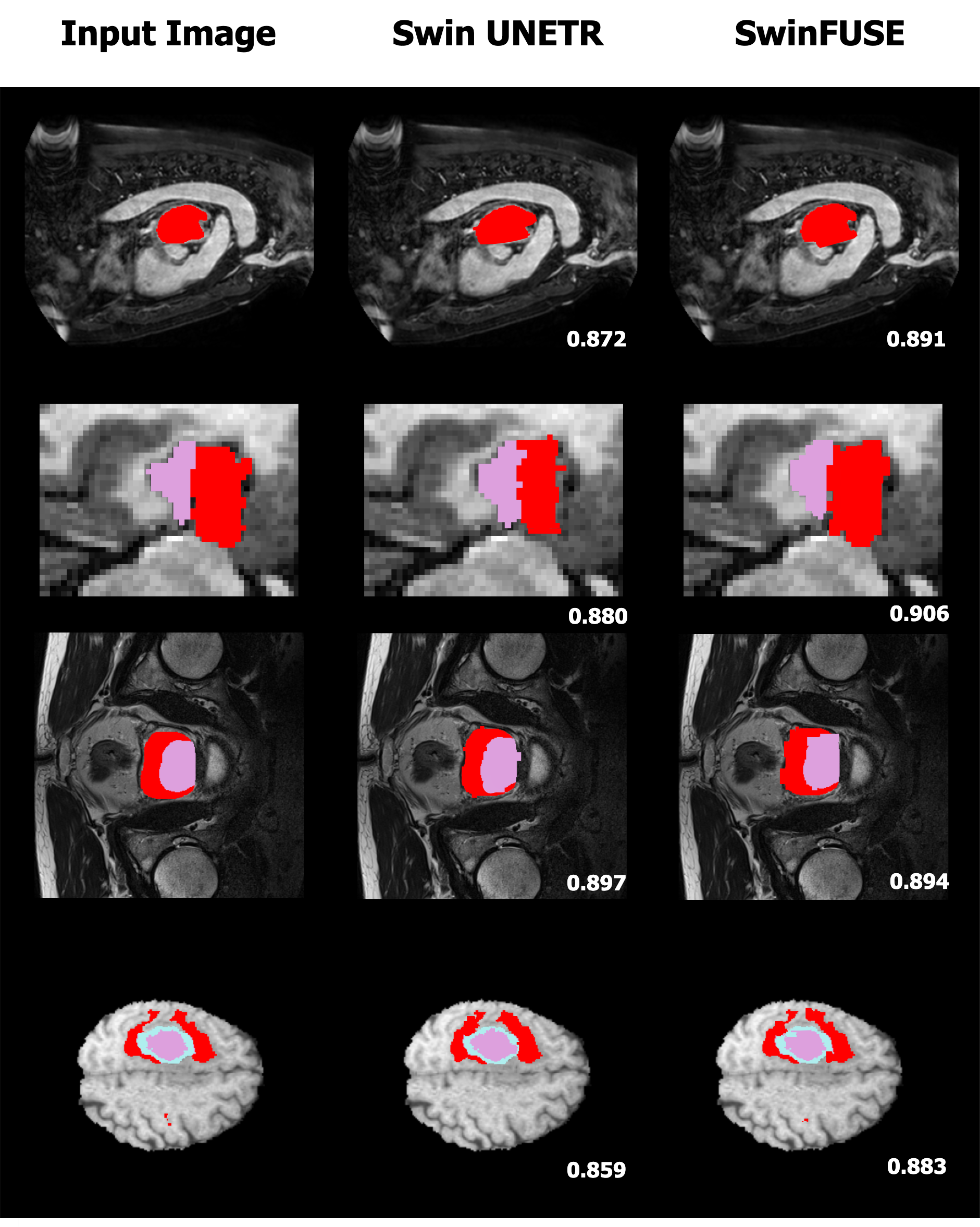} 
    \caption{Qualitative visualizations of Swin UNETR and our proposed method. Colored regions correspond to necrotic tumor core (red), peritumoral edematous tissue (pink), and enhancing tumor (blue). Dice scores are also given.}
\label{fig:segs} 
\end{figure}

\vspace{-5px}
\section{Results}
\label{sec:results}

\begin{table*}
    \caption{Average Dice Score across five folds on MSD for two variants of SwinFUSE: with and without pre-training denoted by the presence and absence of (P) respectively.}
    \centering
    \begin{tabular}{lcccccccccc}
    \hline
         Organ & Brain & Heart & Hippocampus & Liver & Lung & Pancreas & Prostate & Colon & \begin{tabular}{@{}c@{}}Hepatic \\ Vessel\end{tabular}  & Spleen \\
    \hline
         nnUNET \cite{isensee2019nnu} & 64.50 & 94.82 & \textbf{89.76} & 86.67 & 72.78 & 68.31 & \textbf{84.14} & 59.43 & 70.14 & \textbf{96.34} \\
         Swin UNETR \cite{hatamizadeh2021swin} & 66.31 & 94.32 & 88.39 & 87.42 & \textbf{76.40} & \textbf{72.91} & 81.51 & 60.35 & 70.75 & 95.79 \\
         \hline
         SwinFUSE & 65.17 & 94.67 & 86.76 & 88.31 & 74.39 & 70.95 & 80.63 & \textbf{60.42} & 69.78 & 94.54 \\
         SwinFUSE (P) & \textbf{66.34} & \textbf{94.91} & 87.91 & \textbf{89.43} & 75.35 & 69.25 & 81.62 & 59.31 & \textbf{71.61} & 96.24 \\
         \hline
    \end{tabular}
    \label{tab:msd}
\end{table*}

\paragraph*{Quantitative}{Average dice scores across five folds for each task in MSD are detailed in Table. \ref{tab:msd} for nnUnet, Swin UNETR, and our model (with and without pre-trained weights). The proposed method outperforms the current SOTA while segmenting the Brain, Heart, Liver, and Hepatic Vessels but, on average, is 1-2\% worse than Swin UNETR. This might be because we reuse the same pre-trained weights to fine-tune SwinFUSE for both CT and MRI tasks, differing from Swin UNETR, which uses pre-trained weights for only CT tasks.}

\paragraph*{Qualitative}{We visualize images from the MSD (heart, hippocampus, prostate, brain) and BraTS21 datasets in Fig. \ref{fig:segs}. Segmentation outputs from SwinFUSE are more concise and perform well in the global context. Moderate improvements are seen for smaller organs like the hippocampus (second row). In contrast, for organs like the brain (last row), we notice that the attention mechanism helps locate the unconnected region in the right hemisphere, which Swim UNETR completely fails to detect. In Fig. \ref{fig:maps} we generate the learned attention weights from which we can see that the DIM learns similarities between CTs and MRIs and uses those to anchor itself. It further uses differentiating aspects between the both to effectively highlight regions before sending the input to the Swin transformer.}
\vspace{-5px}

\begin{table}[]
    \centering
    \caption{Performance on MRI tasks after pre-training on CT organ regions. Dice scores of Swin UNETR (1) and our model (2) are reported in the format 1/2 in each cell}
    \label{tab:perf}
    \begin{tabular}{lcccc}
        \hline
        \multirow{3}{*}{\begin{tabular}{@{}c@{}}Fine- \\tuned\\on\end{tabular}} & \multicolumn{4}{c}{Tested on} \\
        \cmidrule{2-5}
         & Brain & Heart & Prostate & \begin{tabular}{@{}c@{}}Hippo- \\ campus\end{tabular} \\
        \hline
        Liver & 0.22/\textbf{0.47} & 0.31/\textbf{0.52} & 0.19/\textbf{0.32} & 0.18/\textbf{0.33} \\
        Pancreas & 0.21/\textbf{0.29} & 0.27/\textbf{0.34} & 0.20/\textbf{0.40} & 0.16/\textbf{0.26} \\
        Lung & 0.15/\textbf{0.32} & 0.30/\textbf{0.46} & 0.22/\textbf{0.34} & 0.23/\textbf{0.30} \\
        Spleen & 0.25/\textbf{0.32} & 0.21/\textbf{0.39} & 0.26/\textbf{0.44} & 0.25/\textbf{0.48} \\
        Colon & 0.19/\textbf{0.39} & 0.24/\textbf{0.53} & 0.29/\textbf{0.49} & 0.27/\textbf{0.38} \\
        \begin{tabular}{@{}c@{}}Hepatic \\ Vessel\end{tabular} & 0.17/\textbf{0.35} & 0.22/\textbf{0.47} & 0.29/\textbf{0.43} & 0.25/\textbf{0.41} \\
        \hline
    \end{tabular}
\end{table}

\paragraph*{Out-of-distribution}{When we fine-tune SwinFUSE and Swin UNETR on CT organ regions in MSD like Liver, Lung, and Pancreas, and later test on MRI regions, we notice a significant drop in performance for the latter. For example, when fine-tuned on the Liver and tested on the Brain, Swin UNETR's average dice score is 0.22, whereas SwinFUSE's is 0.47. In another instance, fine-tuning on the Pancreas and testing on the Prostate resulted in Swin UNETR achieving a score of 0.16 compared to 0.26 for ours. Multiple other experiments like these are shown in Table. \ref{tab:perf} where we showcase a minimum improvement of 7\% and a maximum of 27\% in dice scores. In conclusion, due to the pre-training of our model on a varied collection of human body compositions and its ability to acquire a versatile representation from data obtained from various institutions, we assert that our model is more suitable for clinical applications than existing single-modality models.}

\vspace{-5px}
\section{Discussion and Conclusion}
\label{sec:conc}

In our research, we have shown noteworthy improvements in the field of medical imaging by using unsupervised pre-training. However, we acknowledge a challenge known as domain shift, where the data used for pre-training differs from the data used for fine-tuning. To address this, we extended the Swin Transformer framework to pre-train SwinFUSE on two distinct medical imaging modalities, CT and MRI. This extension offers three key advantages:

\begin{itemize}
    \item Complementary Feature Representations: By training on both CT and MRI data, SwinFUSE learns diverse feature representations, making it more adaptable and robust.
    \item Domain-Invariance Module: Our DIM helps SwinFUSE adapt to domain shifts by emphasizing important regions.
    \item Remarkable Generalizability: SwinFUSE can perform well on tasks it wasn't initially trained for, making it highly relevant in real-world applications.
\end{itemize}

Our experiments show that our approach performs slightly worse than single-modality models on in-distribution tasks but significantly outperforms them on out-of-distribution modalities, highlighting its practical applicability. Quantitatively, our method surpassed state-of-the-art models in segmenting the Brain, Heart, Liver, and Hepatic Vessels in the MSD dataset. We emphasize that our diverse pre-training data and versatile representations make SwinFUSE more suitable for clinical use than single-modality models.

We've demonstrated the advantages of pre-training SwinFUSE using our domain-invariance module and its superior performance on various tasks. We've also highlighted the potential for future research in addressing domain gaps and applying our framework to other medical imaging modalities like PET and X-rays. This work represents a significant advancement in the efficiency and accuracy of medical image analysis, with promising prospects for future research.

\section{Compliance with ethical standards}
\label{sec:ethics}

This research study was conducted retrospectively using human subject data available in open access by RSNA ASNR-MICCAI BraTS 2021, SynthRAD 2023, and Medical Segmentation Decathlon challenges. Ethical approval was
not required, as confirmed by the license attached to the
open access data.

\section{Acknowledgments}
\label{sec:acknowledgments}

We thank IHub-Data, International Institute of Information Technology, Hyderabad, for their support.




\bibliographystyle{IEEEbib}
\bibliography{refs}
\vfill

\end{document}